\documentclass{article}

 \usepackage[preprint,nonatbib]{neurips_2019}

\usepackage[utf8]{inputenc} 
\usepackage[T1]{fontenc}    
\usepackage{hyperref}       
\usepackage{url}            
\usepackage{booktabs}       
\usepackage{amsfonts}       
\usepackage{nicefrac}       
\usepackage{microtype}      
\usepackage{enumitem}
\usepackage{outlines}
\usepackage{xspace}
\usepackage{listings}
\usepackage{color}
\usepackage{colortbl}
\usepackage{graphicx}
\usepackage{amsmath, amsfonts}
\usepackage{subcaption}
\usepackage{algorithm}      
\usepackage{algorithmic}    
\usepackage{url}
\usepackage[greek,english]{babel}
\usepackage[utf8]{inputenc}
\usepackage{subcaption}
\usepackage{graphicx} 
\usepackage{verbatim}
\usepackage{pbox}
\usepackage{multirow}

\usepackage[compact, small]{titlesec}

\usepackage{times}

\usepackage{balance}

\usepackage{dsfont}

\usepackage{authblk}

\usepackage[labelfont=bf, textfont=rm, font=small, belowskip=0pt]{caption}

\usepackage{hyperref}
\definecolor{darkred}{rgb}{0.7,0,0}
\definecolor{darkgreen}{rgb}{0,0.5,0}
\hypersetup{colorlinks=true,
        linkcolor=darkred,
        citecolor=darkgreen}
\usepackage{listings}

\def\compactify{\itemsep=0pt \topsep=0pt \partopsep=0pt \parsep=0pt}
\let\latexusecounter=\usecounter

\newenvironment{CompactEnumerate}
  {\def\usecounter{\setlength{\leftmargin}{1.2em}\compactify\latexusecounter}
   \begin{enumerate}}
  {\end{enumerate}\let\usecounter=\latexusecounter}

\newcommand{\name}{Placeto\xspace}
\newcommand{\inception}{Inception-V3}
\newcommand{\nmt}{NMT}
\newcommand{\nasnet}{NASNet}

\newcommand{\kilo}{K}
\newcommand{\rnn}{RNN}
\newcommand{\computational}{computation\xspace}

\title{\name: Learning Generalizable Device Placement Algorithms for Distributed Machine Learning}

\author{
  \textbf{\;\;\;\;\;\;\;\;\;\;} \textbf{Ravichandra Addanki}, \textbf{Shaileshh Bojja Venkatakrishnan}, \textbf{Shreyan Gupta}, \textbf{\;\;\;\;\;\;\;\;\;\;} \textbf{Hongzi Mao}, \textbf{Mohammad Alizadeh}\\
  MIT Computer Science and Artificial Intelligence Laboratory \\
  \texttt{\{addanki, bjjvnkt, shreyang, hongzi, alizadeh\}@mit.edu} \\
}

\newcommand{\eg}{{\em e.g., }}

\begin{document}

\maketitle

\vspace{-0.3cm}
\begin{abstract}

%
%

%
We present \name, a reinforcement learning (RL) approach to
efficiently find device placements for distributed neural network training.
Unlike prior approaches that only find a device placement for a specific \computational graph, \name can learn generalizable device placement policies that can be applied to {\em any} graph.
We propose two key ideas in our approach:
(1) we represent the policy as performing iterative placement improvements, rather than outputting a placement in one shot;
(2) we use graph embeddings to capture relevant information about the  structure of the \computational graph, without relying on node labels for indexing.
These ideas allow \name to train efficiently and generalize
to unseen graphs.
Our experiments show that \name requires up to $6.1\times$
fewer training steps to find placements that are on par with or
better than the best placements found by prior approaches. 
%
Moreover, \name is able to learn a generalizable placement policy for any given family of graphs, which can then be used without any retraining to predict optimized placements for unseen graphs from the same family.
This eliminates the large overhead incurred by prior RL approaches whose lack of generalizability necessitates re-training from scratch every time a new graph is to be placed.

\end{abstract}
\section{Introduction \& Related Work}
\label{s:intro}
\vspace{-.25cm}

The computational requirements for training neural networks have steadily increased in recent years. As a result, a growing number of applications~\cite{massively_parallel_deep_rl, alphagozero} use distributed training environments in which a neural network is split across multiple GPU and CPU devices. A key challenge for distributed training is how to split a large model across multiple heterogeneous devices to achieve the fastest possible training speed. 
Today device placement is typically left to human experts, but determining an optimal device placement can be very challenging, particularly as neural networks grow in complexity (e.g., networks with many interconnected branches) or approach device memory limits. In shared clusters, the task is made even more challenging due to interference and variability caused by other applications.

Motivated by these challenges, a recent line of work~\cite{grl1, grl2, spotlight} has proposed an automated approach to device placement based on reinforcement learning (RL). In this approach, a neural network policy is trained to optimize the device placement through repeated trials. For example,  Mirhoseini et al.~\cite{grl1} use a recurrent neural network (RNN) to process a \computational graph and predict a placement for each operation. They show that the RNN, trained to minimize computation time, produces device placements that outperform both human experts and graph partitioning heuristics such as Scotch~\cite{scotch}. Subsequent work~\cite{grl2} improved the scalability of this approach with a hierarchical model 
and explored more sophisticated policy optimization techniques~\cite{spotlight}.

Although RL-based device placement is promising, existing approaches have a key drawback: they require significant amount of re-training to find a good placement for each \computational graph. For example, Mirhoseini et al.~\cite{grl1} report 12 to 27 hours of training time to find the best device placement for several vision and natural language models; more recently, the same authors report 12.5 GPU-hours of training to find a placement for a neural machine translation (NMT) model~\cite{grl2}. While this overhead may be acceptable in some scenarios (e.g., training a stable model on large amounts of data), it is undesirable in many cases. For example, high device placement overhead is problematic during model development, which can require many ad-hoc model explorations. Also, in a shared, non-stationary environment, it is important to make a placement decision quickly, before the underlying environment changes. 

Existing methods have high overhead because they do not learn {\em generalizable} device placement policies. Instead they optimize the device placement for a {\em single} \computational graph. Indeed, the training process in these methods can be thought of as a search for a good placement for one \computational graph, rather than a search for a good placement {\em policy} for a class of \computational graphs. Therefore, for a new \computational graph, these methods must train the policy network from scratch. Nothing learned from previous  graphs carries over to new graphs, neither to improve placement decisions nor to speed up the search for a good placement.



In this paper, we present \name, a reinforcement learning (RL) approach to learn an efficient algorithm for device placement for a given family of \computational graphs.
Unlike prior work, \name is able to transfer a learned placement policy to unseen \computational graphs from the same family without requiring any retraining.

\name incorporates two key ideas to improve training efficiency and generalizability.
First, it models the device placement task as finding a sequence of {\em iterative placement improvements}.
Specifically, \name's policy network takes as input a current placement for a \computational graph, and one of its node, and it outputs a device for that node. By applying this policy sequentially to all nodes, \name is able to iteratively optimize the placement. 
This placement improvement policy, operating on an explicitly-provided input placement, is simpler to learn than a policy representation that must output a final placement for the entire graph in one step. 

\name's second idea is a neural network architecture that uses {\em graph embeddings}~\cite{graph_emb_survey,graph_cnn_1,graph_cnn_2} to encode the \computational graph structure in the placement policy. Unlike prior RNN-based approaches, \name's neural network policy does not depend on the sequential order of nodes or an arbitrary labeling of the graph (e.g., to encode adjacency information). Instead it naturally captures graph structure (e.g., parent-child relationships) via iterative message passing computations performed on the graph.


Our experiments show that \name learns 
placement policies that outperform the RNN-based 
approach over three neural network models: 
Inception-V3~\cite{inceptionv3}, NASNet~\cite{nasnet} and NMT~\cite{nmt}.
For example, on the NMT model \name finds a placement that runs $16.5\%$ faster than the RNN-based approach.
Moreover, it also learns these placement policies substantially faster,  with up to 6.1$\times$ fewer placement evaluations, than the RNN approach. 
Given any family of graphs \name learns a generalizable placement policy, that can then be used to predict optimized placements for unseen graphs from the same family without any re-training.
This avoids the large overheads incurred by \rnn-based approaches which must repeat the training from scratch every time a new graph is to be placed.

%


Concurrently with this work, Paliwal et al.~\cite{regal} propose using graph embeddings to learn a generalizable policy for device placement and schedule optimization. 
However, their approach does not involve optimizing placements directly; instead a complex genetic search algorithm needs to be run for several thousands of iterations everytime placement for a new graph is to be optimized~\cite{regal}. 
This incurs a large penalty of evaluating thousands of placements and schedules, rendering the generalizability of the learned policy ineffective.

 
\section{Learning Method} 
\label{s:design}
\vspace*{-.2cm}
The \computational graph of a neural network can be modeled as a graph $G(V,E)$, where $V$ denotes the atomic computational operations (also referred to as ``ops'') in the neural network, and $E$ is the set of data communication edges.  
Each op $v\in V$ performs a specific computational function
(e.g., convolution) on input tensors that it receives from its parent ops.
%
For a set of devices $\mathcal{D}=\{d_1,\ldots,d_m\}$, a {\em placement} for $G$ is a mapping $\pi:V\rightarrow D$ that assigns a device to each op.
The goal of device placement is to find a placement $\pi$ that minimizes $\rho(G, \pi)$, the duration of $G$'s execution when its ops are placed according to $\pi$. 
To reduce the number of placement actions, we partition ops into predetermined {\em groups}
and place ops from the same group on the same device, similar to Mirhoseini et.al.~\cite{grl2}.
For ease of notation, henceforth we will use $G(V,E)$ to denote the graph of op groups.
Here $V$ is the set of op groups and $E$ is set of data communication edges between op groups. 
An edge is drawn between two op groups if there exists a pair of ops, from the respective op groups, that have an edge between them in the neural network.

%
\begin{figure*}[t]
  \centering
  \includegraphics[width=\columnwidth]{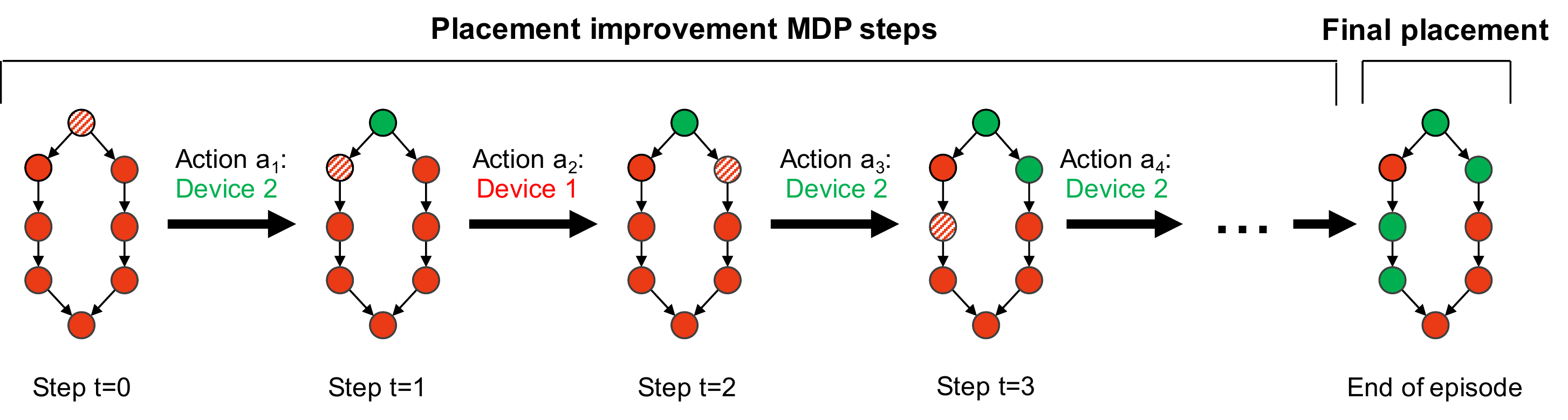}
  \caption{MDP structure of \name's device placement task. At each step, \name updates a placement for a node (shaded) in the \computational graph. These incremental improvements amount to the final placement at the end of an MDP episode.}
  \label{f:mdp}
  \vspace{-0.5cm}
\end{figure*}

%
%

\name finds an efficient placement for a given input \computational graph, by executing an iterative {\em placement improvement policy} on the graph. 
The policy is learned using RL over \computational graphs that are structurally similar (i.e., coming from the same underlying probability distribution) as the input graph. 
In the following we present the key ideas of this learning procedure: the Markov decision process (MDP) formalism in \S\ref{sec: MDP formulation}, graph embedding and the neural network architecture for encoding the placement policy in \S\ref{sec: graph embedding}, and the training/testing methodology in \S\ref{sec: training and testing}. 
We refer the reader to~\cite{sutton_rl_book} for a primer on RL. 

\subsection{MDP Formulation}
\label{sec: MDP formulation}
%
Let $\mathcal{G}$ be a family of \computational graphs, for which we seek to learn an effective placement policy. 
We consider an MDP where a state observation $s$ comprises of a graph $G(V,E)\in \mathcal{G}$ with the following features on each node $v\in V$: (1) estimated run time of $v$, (2) total size of tensors output by $v$, (3) the current device placement of $v$, (4) a flag indicating whether $v$ has been ``visited'' before, and (5) a flag indicating whether $v$ is the ``current'' node for which the placement has to be updated. 
At the initial state $s_0$ for a graph $G(V,E)$, the nodes are assigned to devices arbitrarily, the visit flags are all $0$, and an arbitrary node is selected as the current node. 


At a step $t$ in the MDP, the agent selects an action to update the placement for the current node $v$ in state $s_t$.
The MDP then transitions to a new state $s_{t+1}$ in which $v$ is marked as visited, and an unvisited node is selected as the new current node. 
The episode ends in $|V|$ steps when the placements for all the nodes have been updated.  
This procedure has been illustrated for an example graph to be placed over two devices, in Figure~\ref{f:mdp}. 


We consider two approaches for assigning rewards in the MDP: (1) assigning a zero reward at each intermediate step in the MDP, and a reward equal to the negative run time of the final placement at the terminal step; (2) assigning an intermediate reward of $r_t = \rho(s_{t+1}) - \rho(s_t)$ at the $t$-th round for each $t=0,1,\ldots,|V|-1$, where $\rho(s)$ is the execution time of placement $s$. 
Intermediate rewards can help improve credit assignment in long training episodes and reduce variance of the policy gradient estimates~\cite{hindsight_q_learning, 
reward_shaping, sutton_rl_book}. However, training with intermediate rewards is more expensive, as it must determine the computation time for a placement at each step as opposed to once per episode.
We contrast the benefits of either reward design through evaluations in Appendix~\ref{s:intermediate-rewards}.  
To find a valid placement that fits without exceeding the memory limit on devices, we include a penalty in the reward proportional to the peak memory utilization if it crosses a certain threshold $M$ (details in Appendix~\ref{s:training_details}).

\begin{figure*}[t]
  \centering
  \includegraphics[width=\columnwidth]{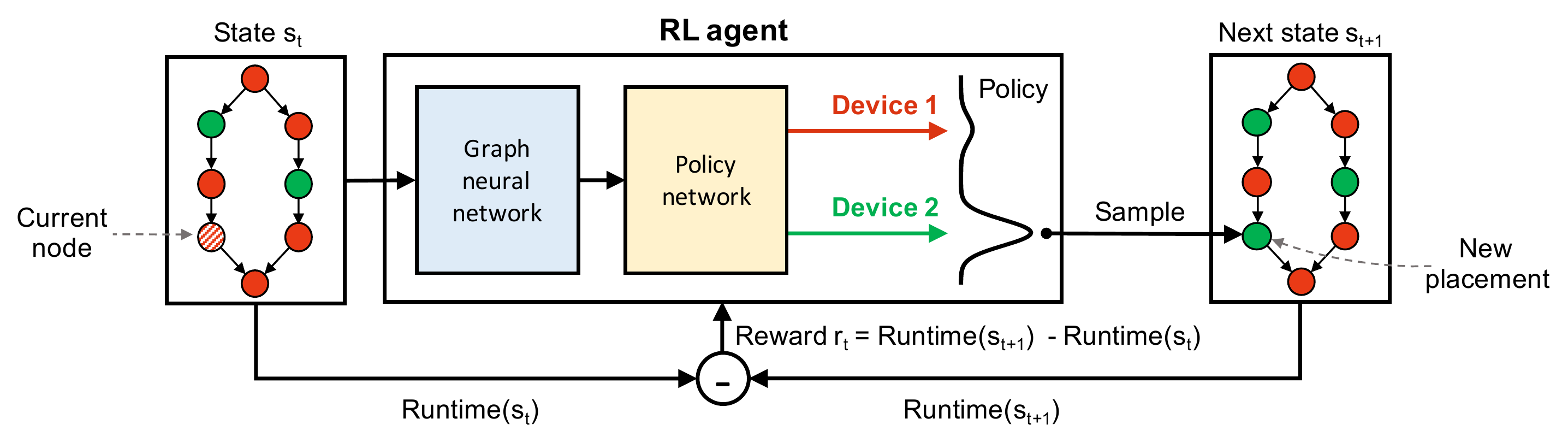}
  \caption{\name's RL framework for device placement. The state input to the agent is represented as a DAG with features (such as computation types, current placement) attached to each node. The agent uses a graph neural network to parse the input and uses a policy network to output a probability distribution over devices for the current node. The incremental reward is the difference between runtimes of consecutive placement plans. }
  \label{f:rl}
\end{figure*}

\begin{figure*}[t]
  \centering
  \includegraphics[width=\columnwidth]{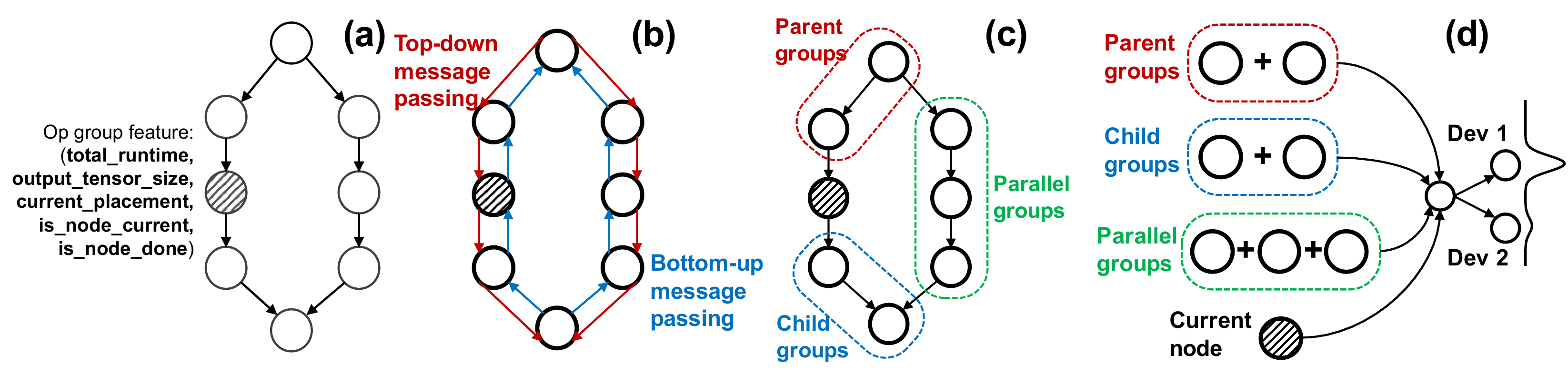}
  \caption{\name's graph embedding approach.
  It maps raw features associated with each op group to the device placement action.
  (a) Example \computational graph of op groups. The shaded node is taking the current placement action.
  (b) Two-way message passing scheme applied to all nodes in the graph. 
  (c) Partitioning the message passed (denoted as bold) op groups.
  (d) Taking a placement action on two candidate devices for the current op group. \vspace{-0.7cm} }
  \label{f:graph_embedding}
\end{figure*}

\subsection{Policy Network Architecture}
\label{sec: graph embedding}
%
\name learns effective placement policies by directly parametrizing the MDP policy using a neural network, which is then trained using a standard policy-gradient algorithm~\cite{Williams92simplestatistical}. 
At each step $t$ of the MDP, the policy network takes the graph configuration in state $s_t$ as input, and outputs an updated placement for the $t$-th node.  
However, to compute this placement action using a neural network, we need to first encode the graph-structured information of the state as a real-valued vector.   
\name achieves this vectorization via a {\em graph embedding} procedure, that is implemented using a specialized graph neural network and  learned jointly with the policy.
  Figure~\ref{f:rl} summarizes how node placements are updated during each round of an RL episode. 
Next, we describe \name's graph neural network.  

\noindent {\bf Graph embedding.}
Recent works~\cite{graph_emb_survey, graph_cnn_1, graph_cnn_2, decima} have proposed graph embedding techniques that have been shown to achieve state-of-the-art performance on a variety of graph tasks, such as node classification, link prediction, job scheduling etc.
Moreover, the embedding produced by these methods are such that they can generalize (and scale) to unseen graphs. 
Inspired by this line of work, in \name we present a graph embedding architecture for processing the raw features associated with each node in the \computational graph. 
Our embedding approach is customized for the placement problem and has the following three steps~(Figure~\ref{f:graph_embedding}): 

\begin{CompactEnumerate}
\item \emph{Computing per-group attributes}~(Figure~\ref{f:graph_embedding}{\color{darkred} a}).
As raw features for each op group, we use the total execution time of ops in the group, total size of their output tensors, a one-hot encoding of the device (e.g., device 1 or device 2) that the group is currently placed on, a binary flag indicating whether the current placement action is for this group, and a binary encoding of whether a placement action has already been made for the group.
We collect the runtime of each op on each device from on-device measurements (we refer to Appendix \ref{s:appendix} for details).
\item \emph{Local neighborhood summarization}~(Figure~\ref{f:graph_embedding}{\color{darkred} b}).
Using the raw features on each node, we perform a sequence of message passing steps~\cite{graph_cnn_1, graph_cnn_2} to aggregate neighborhood information for each node.
Letting $\mathbf{x}_{v}$ denote the features of op group $v$,
  the message passing updates take the form
  $\mathbf{x}_{v} \leftarrow g(\sum_{u \in \xi(v)}f(\mathbf{x}_{u}))$,
  where $\xi(v)$ is the set of neighbors of $v$, and $f,g$ are multilayer perceptrons
  with trainable parameters.
  We construct two directions (top-down from root groups and bottom-up from leaf groups)
  of message passings with separate parameters.
  The top-down messages summarize information about the subgraph of nodes that can reach $v$, while the bottom-up does so for the subgraph reachable from $v$. 
  %
  The parameters in the transformation functions $f,g$ are shared for message passing steps in each direction, among all nodes.  
  We repeat the message passing updates $k$ times to propagate local structural information across the graph, where $k$ is a hyperparameter. 
  As we show in our experiments~(\S\ref{s:evaluation}), reusing the same message passing function
  everywhere provides a natural way to transfer the learned policy to unseen \computational graphs.
  \item \emph{Pooling summaries}~(Figures~\ref{f:graph_embedding}{\color{darkred} c}
  and~\ref{f:graph_embedding}{\color{darkred} d}).
  After message passing, we aggregate the embeddings computed at each node 
  to create a global summary of the entire graph.
  Specifically, for the node $v$ for which a placement decision has to be made,  
  we perform three separate aggregations:
  on the set $S_\text{parents}(v)$ of nodes that can reach $v$,
  set $S_\text{children}(v)$ of nodes that are reachable by $v$,
  and set $S_\text{parallel}(v)$ of nodes that can neither reach nor be reached by $v$.
  On each set $S_i(v)$, we perform the aggregations
  using $h_i(\sum_{u \in S_i(v)} l_i(\mathbf{x}_{u}))$
  where $\mathbf{x}_{u}$ are the node embeddings and $h_i, l_i$ are
  multilayer perceptrons with trainable parameters as above.
  Finally, node $v$'s embedding and the result from the three aggregations are concatenated as input to the subsequent policy network. 
\end{CompactEnumerate}
The above three steps define an end-to-end policy mapping from raw features associated with each op group to the device placement action.

\subsection{Training}
\label{sec: training and testing}
\name is trained using a standard policy-gradient algorithm~\cite{Williams92simplestatistical}, with a timestep-based baseline~\cite{greensmith} (see Appendix~\ref{s:policy_gradient} for details).
During each training episode, a graph from a set $\mathcal{G}_T$ of training graphs is sampled and used for performing the rollout. 
The neural network design of \name's graph embedding procedure and policy network allows the training parameters to be shared across episodes, regardless of the input graph type or size.   
This allows \name to learn placement policies that generalize well to unseen graphs during testing. 
We present further details on training in \S\ref{s:evaluation}.

\section{Experimental Setup}
\label{s:evaluation}

\vspace{-.1cm}

\subsection{Dataset}
\vspace{-.2cm}
We use Tensorflow to generate a \computational graph given any neural network model, which can then be run to perform one step of stochastic gradient descent on a mini-batch of data.
We evaluate our approach on \computational graphs corresponding to the following three popular deep learning models: (1) \textbf{Inception-V3} \cite{inceptionv3}, a widely used convolutional neural network which has been successfully applied to a large variety of computer vision tasks; 
(2) \textbf{NMT} \cite{nmt}, a language translation model that uses an LSTM based encoder-decoder and attention architecture for natural language translation;
(3) \textbf{NASNet} \cite{nasnet}, a computer vision model designed for image classification. For a more detailed descriptions of these models, we refer to Appendix \ref{s:models}

We also evaluate on three synthetic datasets, each comprising of 32 graphs, spanning a wide range of graph sizes and structures. 
We refer to these datasets as \textit{cifar10}, \textit{ptb} and \textit{nmt}. Graphs from \textit{cifar10} and \textit{ptb} datasets are synthesized using an automatic model design approach called ENAS \cite{enas}. 
The \textit{nmt} dataset is constructed by varying the RNN length and batch size hyperparameters of the  \nmt~model~\cite{nmt}. 
We randomly split these datasets for training and test purposes.
Graphs in \textit{cifar10} and \textit{ptb} datasets are grouped to have about 128 nodes each, whereas graphs from \textit{nmt} have 160 nodes.
Further details on how these datasets are constructed can be found in the Appendix \ref{s:datasets}.

\subsection{Baselines}
\vspace{-.1cm}

We compare \name against the following heuristics and baselines from prior work~\cite{grl1, grl2, spotlight}: \\
(1) \textbf{Single GPU}, where all the ops in a model are placed on the same GPU.
For graphs that can fit on a single device and don't have a significant inherent parallelism in their structure, this baseline can often lead to the fastest placement as it eliminates any cost of communication between devices. \\
(2)  \textbf{Scotch}~\cite{scotch}, a graph-partitioning-based static mapper that takes as input the \computational graph, cost associated with each node, amount of data associated with connecting edges, and then outputs a placement which minimizes communication costs while keeping the load balanced across devices within a specified tolerance. \\
(3)  \textbf{Human expert.} For NMT models, we place each LSTM layer on a separate device as recommended by Wu et al.~\cite{nmt}. We also colocate the attention and softmax layers with the final LSTM layer. Similarly for vision models, we place each parallel branch on a different device. \\
(4) \textbf{\rnn-based approach}~\cite{grl2}, in which the placement problem is posed as finding a mapping from an input sequence of op-groups to its corresponding sequence of optimized device placements. 
An RNN model is used to learn this mapping.
The RNN model has an encoder-decoder architecture with content-based attention mechanism. 
We use an open source implementation from Mirhoseini et.al.~\cite{grl2} available as part of the official Tensorflow repository~\cite{github:tensorflow}. 
We use the included hyperparameter settings and tune them extensively as required.

\subsection{Training Details}

\textbf{Co-location groups.}
%
To decide which set of ops have to be co-located in an op-group, we follow the same strategy as described by Mirhoseini et al.~\cite{grl1} and use the  final grouped graph as input to both \name and the \rnn-based approach.
We found that even after this grouping, there could still be a few operation groups with very small memory and compute costs left over. 
We eliminate such groups by iteratively merging them with their neighbors as detailed in Appendix \ref{s:grouping_merge_strategy}.

\textbf{Simulator.}
%
%
Since it can take a long time to execute placements on real hardware and measure the elapsed time ~\cite{grl2, grl1}, we built a reliable simulator that can quickly predict the runtime of any given placement for a given device configuration. 
We have discussed details about how the simulator works and its accuracy in Appendix \ref{s:simulator}. 
This simulator is used only for training purposes.
All the reported runtime improvements have been obtained by evaluating the learned placements on real hardware, unless explicitly specified otherwise.

Further details on training of \name and the \rnn-based approach are given in the Appendix \ref{s:training_details}. 



\vspace*{-.2cm}
\section{Results} \label{s:results} 
\vspace*{-.2cm}
In this section, we first evaluate the performance of \name and compare it with aforementioned baselines (\S\ref{s:performance}). 
Then we evaluate \name's generalizability compared to the RNN-based approach (\S\ref{s:generalizability}).  
Finally, we provide empirical validation for \name's design choices (\S\ref{sec: deep dive}).
%
\subsection{Performance}
\label{s:performance}
%
Table~\ref{t:perf} summarizes the performance of \name and baseline schemes for the Inception-V3, NMT and NASNet models.
We quantify performance along two axes: (i) runtime of the best placement found, and (ii) time taken to find the best placement, measured in terms of the number of placement evaluations required for the RL-based schemes while training. 

For all considered graphs, \name is able to rival or outperform the
best comparing scheme.
\name also finds optimized placements much faster than the \rnn-based approach.
For Inception on 2 GPUs, \name is able to find a placement that is $7.8\%$ faster than the expert placement. Additionally, it requires about $4.8 \times$ fewer samples than the \rnn-based approach. Similarly, for the \nasnet \,model \name outperforms the \rnn-based approach using up to $4.7 \times$ fewer episodes.

\begin{figure*}[t!]
\centering
\hspace*{-.45cm}
\includegraphics[width=\columnwidth]{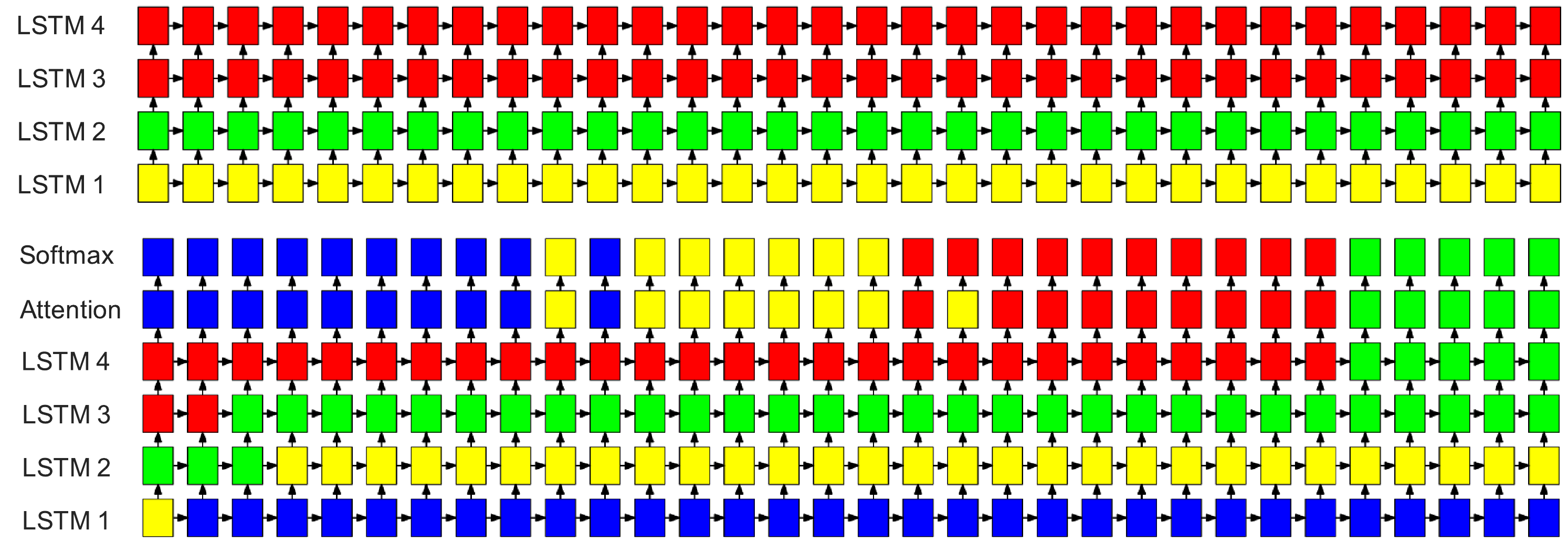}
\caption{
	\small Optimized placement across 4 GPUs for a 4-layered NMT model with attention found by \name.
	The top LSTM layers correspond to encoder and the bottom layers to decoder.
	All the layers are unrolled to a maximum sequence length of 32.
	Each color represents a different GPU.
	This non-trivial placement meets the memory constraints of the GPUs unlike the expert-based placement and the Scotch heuristic, which result in an Out of Memory (OOM) error.
	It also runs 16.5\% faster than the one found by the \rnn-based approach.
	\label{fig:nmt-placement}}
\end{figure*}

%
For the \nmt\, model with 2 GPUs, \name is able to optimize placements to the same extent as the \rnn-based scheme, while using $3.5\times$ fewer samples.
For \nmt~distributed across 4 GPUs, \name finds a non-trivial placement that is $16.5 \%$ faster than the existing baselines. We visualize this placement in Figure \ref{fig:nmt-placement}. 
The expert placement heuristic for NMT fails to meet memory constraints of the GPU devices. This is because in an attempt to maximize parallelism, it places each layer on a different GPU, requiring copying over the outputs of the $i^{th}$ layer to the GPU hosting the $(i+1)^{th}$ layer. These copies have to be retained until they can be fed in as inputs to the co-located gradient operations during the back-propagation phase. This results in a large memory footprint which ultimately leads to an OOM error. On the other hand, \name learns to exploit parallelism and minimize the inter-device communication overheads while remaining within memory constraints of all the devices.
The above results show the advantage of \name's simpler policy representation: it is easier to learn a policy to incrementally improve placements, than to learn a policy that decides placements for all nodes in one shot.
%

\begin{table}[]
\scriptsize
  \centering
  \resizebox{\columnwidth}{!}{%
    \begin{tabular}{|c|cc|ccccc|cc|cc|}
    \hline    
         & \multicolumn{7}{c|}{Placement runtime}  & \multicolumn{2}{c|}{Training time} & \multicolumn{2}{c|}{Improvement} \\
         & \multicolumn{7}{c|}{(sec)}  & \multicolumn{2}{c|}{(\# placements sampled)} & \multicolumn{2}{c|}{} \\
    \hline
        \multirow{2}{*}{Model} & \tiny{CPU} & \tiny{Single} & \multirow{2}{*}{\#GPUs} & \multirow{ 2}{*}{Expert} & \multirow{2}{*}{Scotch} & \multirow{2}{*}\name & \rnn- & \multirow{2}{*}\name & \rnn- & Runtime & Speedup \\
        
        & \tiny{only} & \tiny{GPU} & & & & & based & & based & Reduction & factor \\ 
    \hline    
        \multirow{3}{*}{\inception} \, & \multirow{3}{*}{12.54} & \multirow{3}{*}{1.56} & \multirow{1.5}{*}{2} & \multirow{1.5}{*}{1.28} & \multirow{1.5}{*}{1.54} & \multirow{1.5}{*}{1.18} & \multirow{1.5}{*}{1.17} & \multirow{1.5}{*}{1.6 \kilo} & \multirow{1.5}{*}{7.8 \kilo} & \multirow{1.5}{*}{- 0.85\%} & \multirow{1.5}{*}{\textbf{4.8 $\times$}} \\

         & & & \multirow{2.25}{*}{4} & \multirow{2.25}{*}{1.15} & \multirow{2.25}{*}{1.74} & \multirow{2.25}{*}{1.13} & \multirow{2.25}{*}{1.19} & \multirow{2.25}{*}{5.8 \kilo} & \multirow{2.25}{*}{35.8 \kilo} & \multirow{2.25}{*}{5\%} & \multirow{2.25}{*}{\textbf{6.1 $\times$}} \\ 
         & & & & & & & & &  &  & \\ 

    \hline
        \multirow{3}{*}{\nmt} \, & \multirow{3}{*}{33.5} & \multirow{3}{*}{OOM} & \multirow{1.5}{*}{2} & \multirow{1.5}{*}{OOM} & \multirow{1.5}{*}{OOM} & \multirow{1.5}{*}{2.32} & \multirow{1.5}{*}{2.35} & \multirow{1.5}{*}{20.4 \kilo} & \multirow{1.5}{*}{73 \kilo} & \multirow{1.5}{*}{1.3 \%} & \multirow{1.5}{*}{\textbf{3.5 $\times$}} \\
        
         & & & \multirow{2.25}{*}{4} & \multirow{2.25}{*}{OOM} & \multirow{2.25}{*}{OOM} & \multirow{2.25}{*}{2.63} & \multirow{2.25}{*}{3.15} & \multirow{2.25}{*}{94 \kilo} & \multirow{2.25}{*}{51.7 \kilo} & \multirow{2.25}{*}{\textbf{16.5 \%}} & \multirow{2.25}{*}{0.55 $\times$} \\

        & & & & & & & & &  &  & \\ 

	\hline
	   \multirow{3}{*}{\nasnet} \, & \multirow{3}{*}{37.5} & \multirow{3}{*}{1.28} & \multirow{1.5}{*}{2} & \multirow{1.5}{*}{0.86} & \multirow{1.5}{*}{1.28} & \multirow{1.5}{*}{0.86} & \multirow{1.5}{*}{0.89} & \multirow{1.5}{*}{3.5 \kilo} & \multirow{1.5}{*}{16.3 \kilo} & \multirow{1.5}{*}{3.4\%} & \multirow{1.5}{*}{\textbf{4.7 $\times$}} \\
     
       & & & \multirow{2.25}{*}{4} & \multirow{2.25}{*}{0.84} & \multirow{2.25}{*}{1.22} & \multirow{2.25}{*}{0.74} & \multirow{2.25}{*}{0.76} & \multirow{2.25}{*}{29 \kilo} & \multirow{2.25}{*}{37 \kilo} & \multirow{2.25}{*}{2.6\%} & \multirow{2.25}{*}{\textbf{1.3 $\times$}} \\

      & & & & & & & & &  &  & \\ 
	\hline
    \end{tabular}
} 
\vspace{0.2cm}
\caption{\small Running times of placements found by \name compared with \rnn-based approach~\cite{grl1}, Scotch and human-expert baseline. The number of measurements needed to find the best placements for \name and the RNN-based are also shown (K stands for kilo). Reported runtimes are measured on real hardware. Runtime reductions and speedup factors are calculated with respect to the RNN-based approach. Lower runtimes and lower training times are better. OOM: Out of Memory. For NMT model, the number of LSTM layers is chosen based on the number of GPUs.
	 \label{t:perf}}
 \vspace*{-0.7cm}
\end{table}

\subsection{Generalizability}
\vspace{-.2cm}
\label{s:generalizability}
%
We evaluate generalizability of the learning-based schemes, by training them over a family of graphs, and using the learned policies to predict effective placements for unseen graphs from the same family. 

%
If the placements predicted by a policy are as good as placements found by separate optimizations over the individual test graphs, we conclude that the placement scheme generalizes well. 
Such a policy can then be applied to a wide variety of structurally-similar graphs without requiring re-training.
We consider three family of graphs---{\em nmt, ptb} and {\em cifar10} datasets---for this experiment. 

For each test graph in a dataset, we compare placements generated by the following schemes: 
\textit{(1) \name Zero-Shot.} A \name policy trained over graphs from the dataset, and used to predict placements for the test graph without any further re-training. 
\textit{(2) \name Optimized.} A \name policy trained specifically over the test graph to find an effective placement.  
\textit{(3) Random.} A simple strawman policy that generates placement for each node by sampling from a uniform random distribution. 
%
%
%
%
%
We define \textit{\rnn\, Zero-Shot} and \textit{\rnn\, Optimized} in a similar manner for the \rnn-based approach.

%
\begin{figure*}[t!]
\centering
\includegraphics[width=\columnwidth]{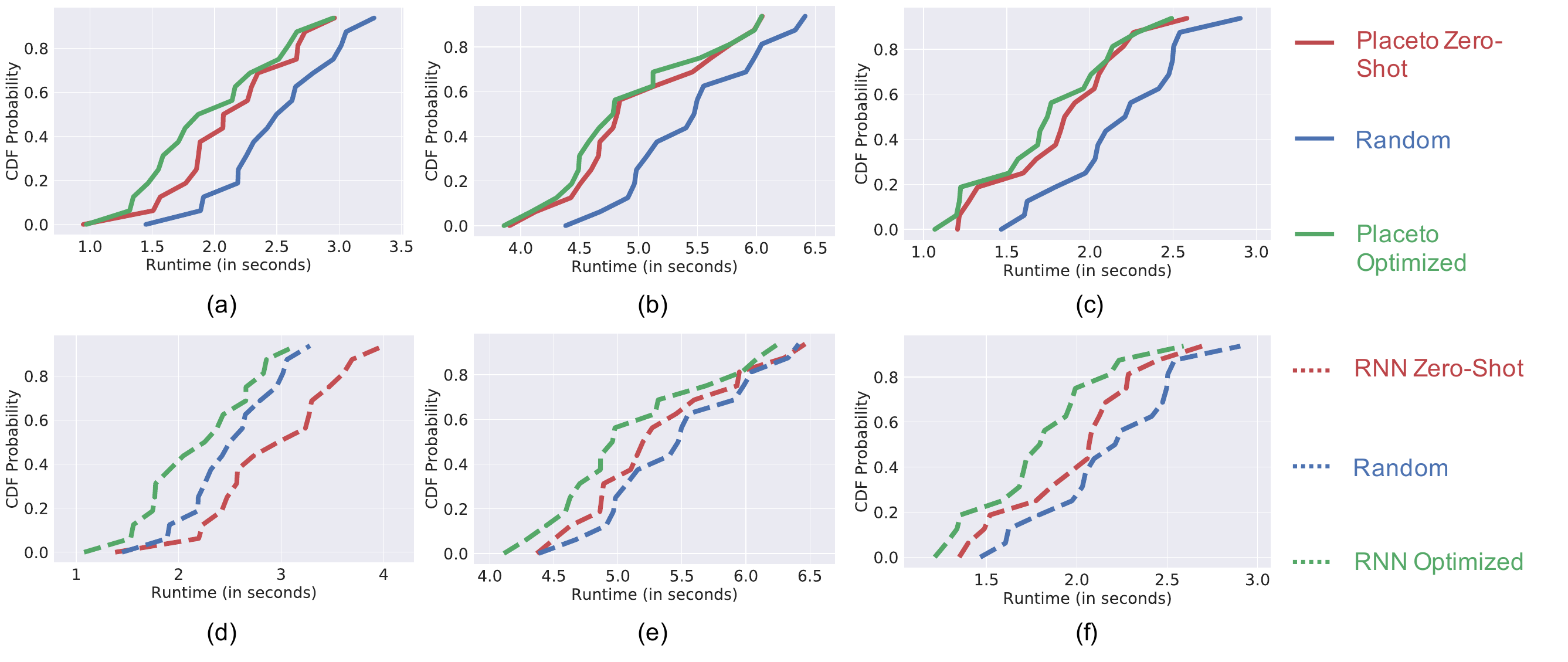}
\vspace{-0.45cm}
\caption{\small CDFs of runtime of placements found by the different schemes for test graphs from (a), (d) \textit{nmt} (b), (e) \textit{ptb} and (c), (f) \textit{cifar10} datasets. Top row of figures ((a), (b), (c)) correspond to \name and bottom row ((d), (e), (f)) to \rnn-based approach. \textit{\name Zero-Shot} performs almost on par with fully optimized schemes like \textit{\name Optimized} and \textit{\rnn\, Optimized} even without any re-training. In contrast, \textit{\rnn\, Zero-Shot} performs much worse and only slightly better than a randomly initialized policy used in \textit{Random} scheme. 
}
\label{fig:generalizability}
\end{figure*}

Figure \ref{fig:generalizability} shows CDFs of runtimes of the placements generated by the above-defined schemes for test graphs from \textit{nmt,\, ptb} and \textit{cifar10} datasets.
We see that the runtimes of the placements generated by \textit{\name Zero-Shot} are very close to those generated by {\it \name Optimized}.
%
Due to \name's generalizability-first design, \textit{\name Zero-Shot} avoids the significant overhead incurred by \textit{\name~Optimized} and \textit{\rnn~Optimized} approaches, which search through several thousands of placements before finding a good one.

Figure \ref{fig:generalizability} also shows that \textit{\rnn\, Zero-Shot} performs significantly worse than \textit{\rnn\, Optimized}. 
In fact, its performance is very similar to that of \textit{Random}.
When trained on a graph, the \rnn-based approach learns a policy to search for an effective placement for that graph. 
However, this learned search strategy is closely tied to the assignment of node indices and the traversal order of the nodes in the graph, which are arbitrary and have a meaning only within the context of that specific graph. 
As a result, the learned policy cannot be applied to graphs with a different structure or even to the same graph using a different assignment of node indices or traversal order.

\subsection{\name Deep Dive}
\label{sec: deep dive}
\vspace{-.3cm}
In this section we evaluate how the node traversal order of a graph during training, affects the policy learned by the different learning schemes. 
We also present an ablation study of \name's policy network architecture.  
In Appendix~\ref{s:intermediate-rewards} we conduct a similar study on the benefits of providing intermediate rewards in \name during training. 


\textbf{Node traversal order.}
Unlike the \rnn-based approach, \name's use of a graph neural network eliminates the need to assign arbitrary indices to nodes while embedding the graph features.
This aids in \name's generalizability, and allows it to learn effective policies that are not tied to the specific node traversal orders seen during training.
To verify this claim, we train \name and the \rnn-based approach on the \inception\, model following one of 64 fixed node traversal orders at each episode.
%
%
%
%
%
We then use the learned policies to predict placements under 64 unseen random node traversal orders for the same model.
With \name, we observe that the predicted placements have runtimes within 5\% of that of the optimized placement on average, with a difference of about 10\% between the fastest and slowest placements.  
%
However, the \rnn-based approach predicts placements that are about 30\% worse on average.
%



\textbf{Alternative policy architectures.}
%
To highlight the role of \name's graph neural network architecture (\S\ref{sec: graph embedding}), we consider the following two alternative policy architectures and compare their generalizability performance against \name's on the {\it nmt} dataset. \\
\textit{(1) Simple aggregator}, in which a feed-forward network is used to aggregate all the node features of the input graph,  which is then fed to another feed-forward network with softmax output units for predicting a placement. This simple aggregator performs very poorly, with its predicted placements on the test dataset about 20\% worse on average  compared to \name. \\
 \textit{(2) Simple partitioner}, in which the node features corresponding to the parent, child and parallel nodes---of the node for which a decision is to be made---are aggregated independently by three different feed-forward networks. 
 Their outputs are then fed to a separate feed-forward network with softmax output units as in the simple aggregator. 
 Note that this is similar to \name's policy architecture (\S\ref{sec: graph embedding}), except for the local neighborhood summarization step (i.e., step 2 in \S\ref{sec: graph embedding}).
 This results in the simple partitioner predicting placements that run 13\% slower on average compared to \name. 
%
Thus, local neighborhood aggregation and pooling summaries from parent, children and parallel nodes are both essential steps for transforming raw node features into generalizable embeddings in \name.


\vspace{-.2cm}
\vspace{-.1cm}
\section{Conclusion}
\label{s:conclusion}
\vspace*{-.3cm}
We presented \name, an RL-based approach for finding device placements
to minimize training time of deep-learning models. 
By structuring the policy decisions as incremental placement
improvement steps, and using graph embeddings to encode
graph structure, \name is able to train efficiently and learns
policies that generalize to unseen graphs.
%
%
\newpage

\bibliographystyle{abbrv}
\small
\bibliography{tfppaper}

\begin{thebibliography}{10}

\bibitem{aws_instance_page}
Ec2 instance types.
\newblock \url{https://aws.amazon.com/ec2/instance-types/}, 2018.
\newblock Accessed: 2018-10-19.

\bibitem{hindsight_q_learning}
M.~Andrychowicz, F.~Wolski, A.~Ray, J.~Schneider, R.~Fong, P.~Welinder,
  B.~McGrew, J.~Tobin, O.~P. Abbeel, and W.~Zaremba.
\newblock Hindsight experience replay.
\newblock In {\em Advances in Neural Information Processing Systems}, pages
  5048--5058, 2017.

\bibitem{graph_emb_survey}
P.~W. Battaglia~et al.
\newblock Relational inductive biases, deep learning, and graph networks.
\newblock {\em arXiv preprint arXiv:1806.01261}, 2018.

\bibitem{graph_cnn_1}
M.~M. Bronstein, J.~Bruna, Y.~LeCun, A.~Szlam, and P.~Vandergheynst.
\newblock Geometric deep learning: going beyond euclidean data.
\newblock {\em IEEE Signal Processing Magazine}, 34(4):18--42, 2017.

\bibitem{spotlight}
Y.~Gao, L.~Chen, and B.~Li.
\newblock Spotlight: Optimizing device placement for training deep neural
  networks.
\newblock In J.~Dy and A.~Krause, editors, {\em Proceedings of the 35th
  International Conference on Machine Learning}, volume~80 of {\em Proceedings
  of Machine Learning Research}, pages 1676--1684, Stockholmsmässan, Stockholm
  Sweden, 10--15 Jul 2018. PMLR.

\bibitem{greensmith}
E.~Greensmith, P.~L. Bartlett, and J.~Baxter.
\newblock Variance reduction techniques for gradient estimates in reinforcement
  learning.
\newblock {\em Journal of Machine Learning Research}, 5(Nov):1471--1530, 2004.

\bibitem{graph_cnn_2}
W.~Hamilton, Z.~Ying, and J.~Leskovec.
\newblock Inductive representation learning on large graphs.
\newblock In {\em Advances in Neural Information Processing Systems}, pages
  1024--1034, 2017.

\bibitem{decima}
H.~Mao, M.~Schwarzkopf, S.~B. Venkatakrishnan, Z.~Meng, and M.~Alizadeh.
\newblock Learning scheduling algorithms for data processing clusters.
\newblock {\em arXiv preprint arXiv:1810.01963}, 2018.

\bibitem{grl2}
A.~Mirhoseini, A.~Goldie, H.~Pham, B.~Steiner, Q.~V. Le, and J.~Dean.
\newblock A hierarchical model for device placement.
\newblock In {\em International Conference on Learning Representations}, 2018.

\bibitem{grl1}
A.~Mirhoseini, H.~Pham, Q.~Le, M.~Norouzi, S.~Bengio, B.~Steiner, Y.~Zhou,
  N.~Kumar, R.~Larsen, and J.~Dean.
\newblock Device placement optimization with reinforcement learning.
\newblock 2017.

\bibitem{massively_parallel_deep_rl}
A.~Nair, P.~Srinivasan, S.~Blackwell, C.~Alcicek, R.~Fearon, A.~De~Maria,
  V.~Panneershelvam, M.~Suleyman, C.~Beattie, S.~Petersen, et~al.
\newblock Massively parallel methods for deep reinforcement learning.
\newblock {\em arXiv preprint arXiv:1507.04296}, 2015.

\bibitem{reward_shaping}
A.~Y. Ng, D.~Harada, and S.~J. Russell.
\newblock Policy invariance under reward transformations: Theory and
  application to reward shaping.
\newblock In {\em Proceedings of the Sixteenth International Conference on
  Machine Learning}, ICML '99, pages 278--287, San Francisco, CA, USA, 1999.
  Morgan Kaufmann Publishers Inc.

\bibitem{github:onnx}
{ONNX Developers}.
\newblock Onnx model zoo, 2018.

\bibitem{regal}
A.~Paliwal, F.~Gimeno, V.~Nair, Y.~Li, M.~Lubin, P.~Kohli, and O.~Vinyals.
\newblock Regal: Transfer learning for fast optimization of computation graphs.
\newblock {\em arXiv preprint arXiv:1905.02494}, 2019.

\bibitem{scotch}
F.~Pellegrini.
\newblock {A parallelisable multi-level banded diffusion scheme for computing
  balanced partitions with smooth boundaries}.
\newblock In T.~P. A.-M.~Kermarrec, L.~Boug{\'e}, editor, {\em {EuroPar}},
  volume 4641 of {\em Lecture Notes in Computer Science}, pages 195--204,
  Rennes, France, Aug. 2007. {Springer}.

\bibitem{enas}
H.~Pham, M.~Y. Guan, B.~Zoph, Q.~V. Le, and J.~Dean.
\newblock Efficient neural architecture search via parameter sharing.
\newblock {\em arXiv preprint arXiv:1802.03268}, 2018.

\bibitem{alphagozero}
D.~Silver, J.~Schrittwieser, K.~Simonyan, I.~Antonoglou, A.~Huang, A.~Guez,
  T.~Hubert, L.~Baker, M.~Lai, A.~Bolton, et~al.
\newblock Mastering the game of go without human knowledge.
\newblock {\em Nature}, 550(7676):354, 2017.

\bibitem{sutton_rl_book}
R.~S. Sutton and A.~G. Barto.
\newblock {\em Introduction to Reinforcement Learning}.
\newblock MIT Press, Cambridge, MA, USA, 1st edition, 1998.

\bibitem{inceptionv3}
C.~Szegedy, V.~Vanhoucke, S.~Ioffe, J.~Shlens, and Z.~Wojna.
\newblock Rethinking the inception architecture for computer vision.
\newblock In {\em Proceedings of the IEEE conference on computer vision and
  pattern recognition}, pages 2818--2826, 2016.

\bibitem{github:tensorflow}
{Tensorflow contributors}.
\newblock Tensorflow official repository, 2017.

\bibitem{wiki:perplexity}
{Wikipedia contributors}.
\newblock Perplexity --- {Wikipedia}{,} the free encyclopedia, 2019.
\newblock [Online; accessed 26-April-2019].

\bibitem{Williams92simplestatistical}
R.~J. Williams.
\newblock Simple statistical gradient-following algorithms for connectionist
  reinforcement learning.
\newblock {\em Machine learning}, 8(3-4):229--256, 1992.

\bibitem{nmt}
Y.~{Wu}, M.~{Schuster}, Z.~{Chen}, Q.~V. {Le}, M.~{Norouzi}, W.~{Macherey},
  M.~{Krikun}, Y.~{Cao}, Q.~{Gao}, K.~{Macherey}, J.~{Klingner}, A.~{Shah},
  M.~{Johnson}, X.~{Liu}, {\L}.~{Kaiser}, S.~{Gouws}, Y.~{Kato}, T.~{Kudo},
  H.~{Kazawa}, K.~{Stevens}, G.~{Kurian}, N.~{Patil}, W.~{Wang}, C.~{Young},
  J.~{Smith}, J.~{Riesa}, A.~{Rudnick}, O.~{Vinyals}, G.~{Corrado},
  M.~{Hughes}, and J.~{Dean}.
\newblock {Google's Neural Machine Translation System: Bridging the Gap between
  Human and Machine Translation}.
\newblock {\em ArXiv e-prints}, 2016.

\bibitem{nasnet}
B.~Zoph, V.~Vasudevan, J.~Shlens, and Q.~V. Le.
\newblock Learning transferable architectures for scalable image recognition.
\newblock {\em arXiv preprint arXiv:1707.07012}, 2(6), 2017.

\end{thebibliography}

\appendix

\newpage
\label{s:appendix}
\noindent
\textbf{\Large Appendices}

\section{Implementation Details}
\label{s:implementation}

\subsection{REINFORCE Algorithm}
\label{s:policy_gradient}

\name is trained using the REINFORCE policy-gradient algorithm~\cite{Williams92simplestatistical}, in which a Monte-Carlo estimate of the gradient is used for updating policy parameters. 
During each training episode, a graph $G$ is sampled from the set of training graphs $\mathcal{G}_T$ (see \S\ref{sec: MDP formulation}) and a rollout $(s_t,a_t,r_t)_{t=0}^{N-1}$ is performed on $G$ using the current policy $\pi_\theta$.  
Here $s_t, a_t, r_t$ refer to the state, action and reward at time-step $t$ respectively, and $\theta$ is the parameter vector encoding the policy. 
At the end of each episode, the policy parameter $\theta$ is updated as 
\begin{align}
\theta \leftarrow \theta + \eta \sum_{i=0}^{N-1} \nabla_\theta \log \pi_\theta(a_i | s_i) \left(\sum_{i'=i}^{N-1} r_{i'} - b_i \right), \label{eq: reinforce}
\end{align}
where $b_i$ is a baseline for reducing variance of the estimate, and $\eta$ is a learning rate hyperparameter.  
\name uses a time-based baseline in which $b_i$ is computed as the average of cumulative rewards $\sum_{i'=i}^{N-1} r_t$ at time-step $i$ over multiple independent rollouts of graph $G$ using the current policy $\pi_\theta$.  
Intuitively, the update rule in Equation~\eqref{eq: reinforce} shifts $\theta$ such that the probability of making ``good" placement actions (i.e., actions for which the cumulative rewards are higher than the average reward) is increased and vice-versa.  
Thus over the course of training, \name gradually learns placement policies for which the overall running time of graphs, coming from the same distribution as $\mathcal{G}_T$, are minimized. 

\subsection{Models}
\label{s:models}
We evaluate our approach on the following popular deep-learning models from Computer Vision and NLP tasks:

%

\begin{enumerate}
    \item \textbf{Inception-V3} \cite{inceptionv3} is a widely used convolutional neural network which has been successfully applied to a large variety of computer vision tasks. Its network consists of a chain of blocks, each of which has  multiple branches made up of convolutional and pooling operations. While these branches from a block can be executed in parallel, each block has a sequential data dependency on its predecessor. The network's input is a batch of $64$ images each with dimension $299\times299\times3$. Its computational graph in tensorflow has $3002$ operations.

    \item \textbf{NMT} \cite{nmt} Neural Machine Translation with attention is a language translation model that uses an LSTM based encoder-decoder architecture to translate a source sequence into a target sequence. When its computational graph is unrolled to handle input sequences of length up to $32$, the memory footprint to hold the LSTM hidden states can be large, potentiating the use of model parallelism. We consider 2-layer as well as 4-layer versions depending on the number of GPUs available for placement. Their computational graphs in tensorflow have $6361$ and $10812$ operations respectively. We use a batch size of $128$.

    \item \textbf{Nasnet} \cite{nasnet} is a computer vision model designed for image classification. Its network consists of a series of cells each of which has multiple branches of computations that are finally reduced at the end to form input for the next cell. It's computational graph consists of $12942$ operations. We use a batch size of $64$.
\end{enumerate}

Prior works \cite{grl1, grl2, spotlight} report significant possibilities of improvements in runtimes for several of the above models when placed over multiple GPUs.


\subsection{Datasets}
\label{s:datasets}

We evaluate the generalizability of each placement scheme by measuring how well it transfers a placement policy learned using the graphs from a training dataset to  unseen graphs from a test dataset.

To our knowledge, there is no available compilation of tensorflow models that is suitable to be used as a training dataset for the device placement problem. For example, one of the most popular tensorflow model collection called ONNX \cite{github:onnx} has only a handful of models and most of them do not have any inherent model parallelism in their computational graph structure.

To overcome this difficulty, we use an automatic model design approach called ENAS \cite{enas} to generate a variety of neural network architectures of different shapes and sizes.
ENAS uses a Reinforcement learning-based controller to discover neural network architectures by searching for an optimal subgraph within a larger graph. It is trained to maximize expected reward on a validation set.

We use the classification accuracy on CIFAR-10 dataset as a reward signal to the controller so that over the course of its training, it generates several neural network architectures which are designed to achieve high accuracy on the CIFAR-10 image classification task.

We randomly sample from these architectures to form a family of $N$ tensorflow graphs which we refer to as the \textit{cifar-10} dataset. Furthermore, for each of these graphs, batch size is chosen by uniformly sampling from the interval, $bs_{low}$ to $bs_{high}$ creating a range of memory requirements for the resulting graphs. We use a fraction $f$ of these graphs for training and the remaining for testing.

Similar to the \textit{cifar-10} dataset, we use the inverse of validation perplexity~\cite{wiki:perplexity} on Penn Treebank dataset as a reward signal to generate a class of tensorflow graphs suitable for language modeling task which we refer to as the \textit{ptb} dataset. Furthermore, we also vary the number of unrolled steps $L$ for the recurrent cell by sampling uniformly from $L_{low}$ to $L_{high}$.

In addition to the above two datasets created using the ENAS method, we create a third dataset made of graphs based on the NMT model which we refer to as the \textit{nmt} dataset. We generate $N$ different variations of the 2-layer NMT model by sampling the number of unrolled steps, $L$ from $L_{low}$ to $L_{high}$ and batch size from $bs_{low}$ to $bs_{high}$. This creates a range of complex graphs based on the common encoder-decoder with attention structure with a wide range of memory requirements.

For our experiments, we use the following settings: 
$N = 32$,\, $f = 0.5$,\, $bs_{low} = 240$\,, $bs_{high} = 360$ for \textit{cifar10} graph dataset,
$N = 32$,\, $f = 0.5$,\, $bs_{low} = 1536$,\, $bs_{high} = 3072$,\, $L_{low} = 25$,\, $L_{high} = 40$\, for \textit{ptb} dataset and
$N = 32$,\, $f = 0.5$,\, $bs_{low} = 64$,\, $bs_{high} = 128$,\, $L_{low} = 16$,\, $L_{high} = 32$\, for \textit{nmt} dataset.

We visualize some samples graphs from \textit{cifar-10} and \textit{ptb} datasets in Figures \ref{f:cifar10-dataset} and \ref{f:ptb-dataset}

\newcommand{\vertspace}{1cm}

\begin{figure*}
\centering

\begin{subfigure}{\columnwidth}
\centering
    \includegraphics[width=\columnwidth]{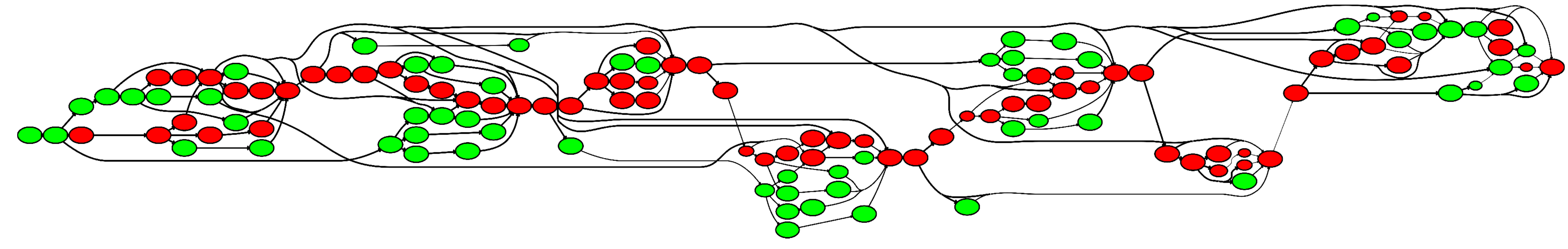}
\end{subfigure}

\vspace{\vertspace}

\begin{subfigure}{\columnwidth}
\centering
    \includegraphics[width=\columnwidth]{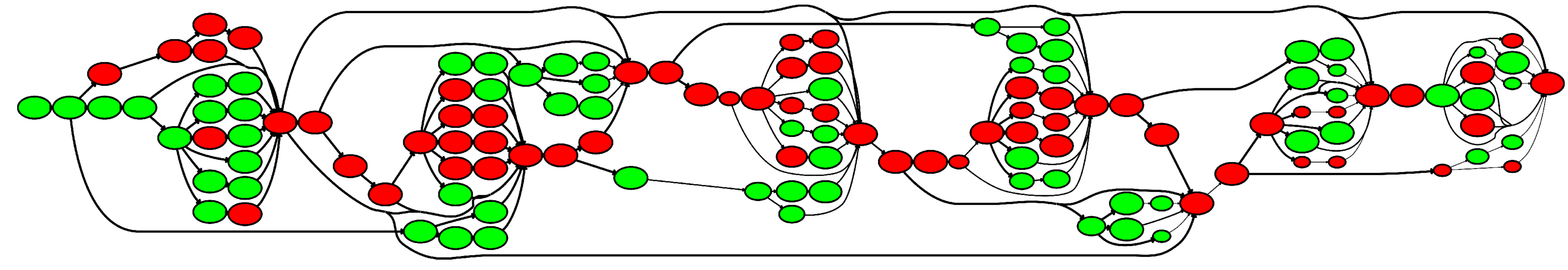}
\end{subfigure}

\vspace{\vertspace}

\begin{subfigure}{\columnwidth}
\centering
    \includegraphics[width=\columnwidth]{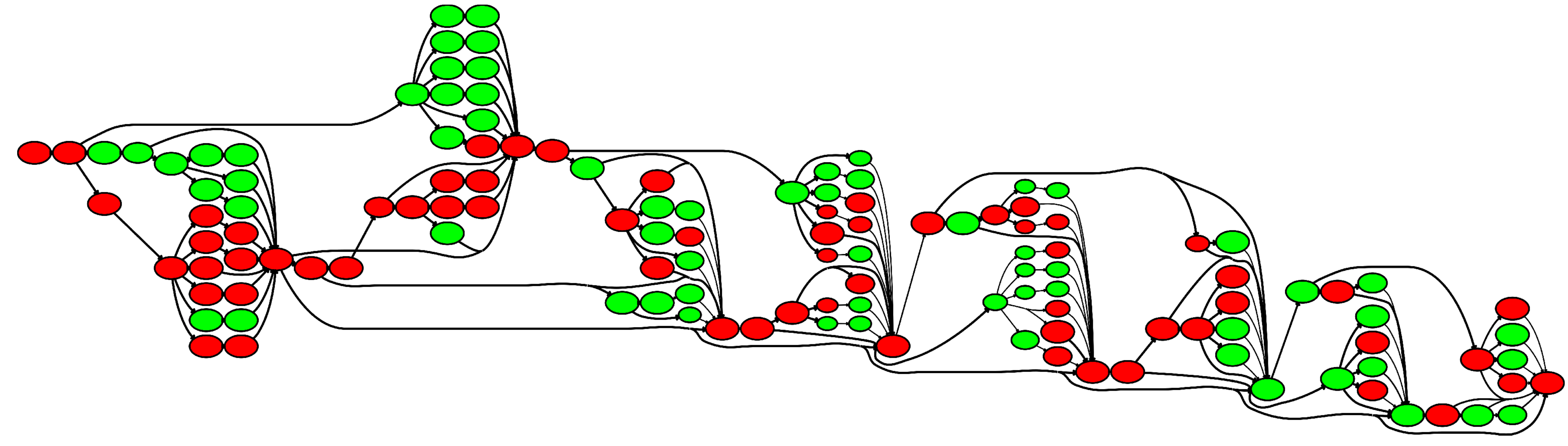}
\end{subfigure}


\begin{subfigure}{\columnwidth}
\centering
    \includegraphics[width=\columnwidth]{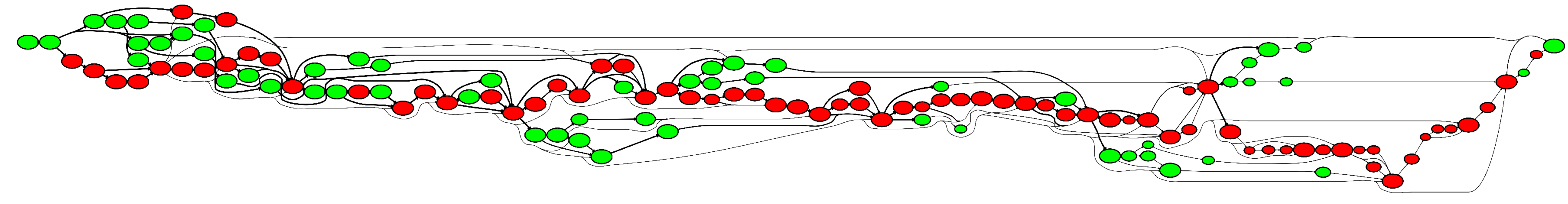}
\end{subfigure}

\caption{Sample graphs from \textit{cifar-10} dataset. Each color represents a different GPU in the optimized placement. Size of the node indicates its compute cost and the edge thickness visualizes the communication cost. The above graphs exhibit a wide range of structure and connectivity.
	\label{f:cifar10-dataset}
}

\includegraphics[width=\linewidth]{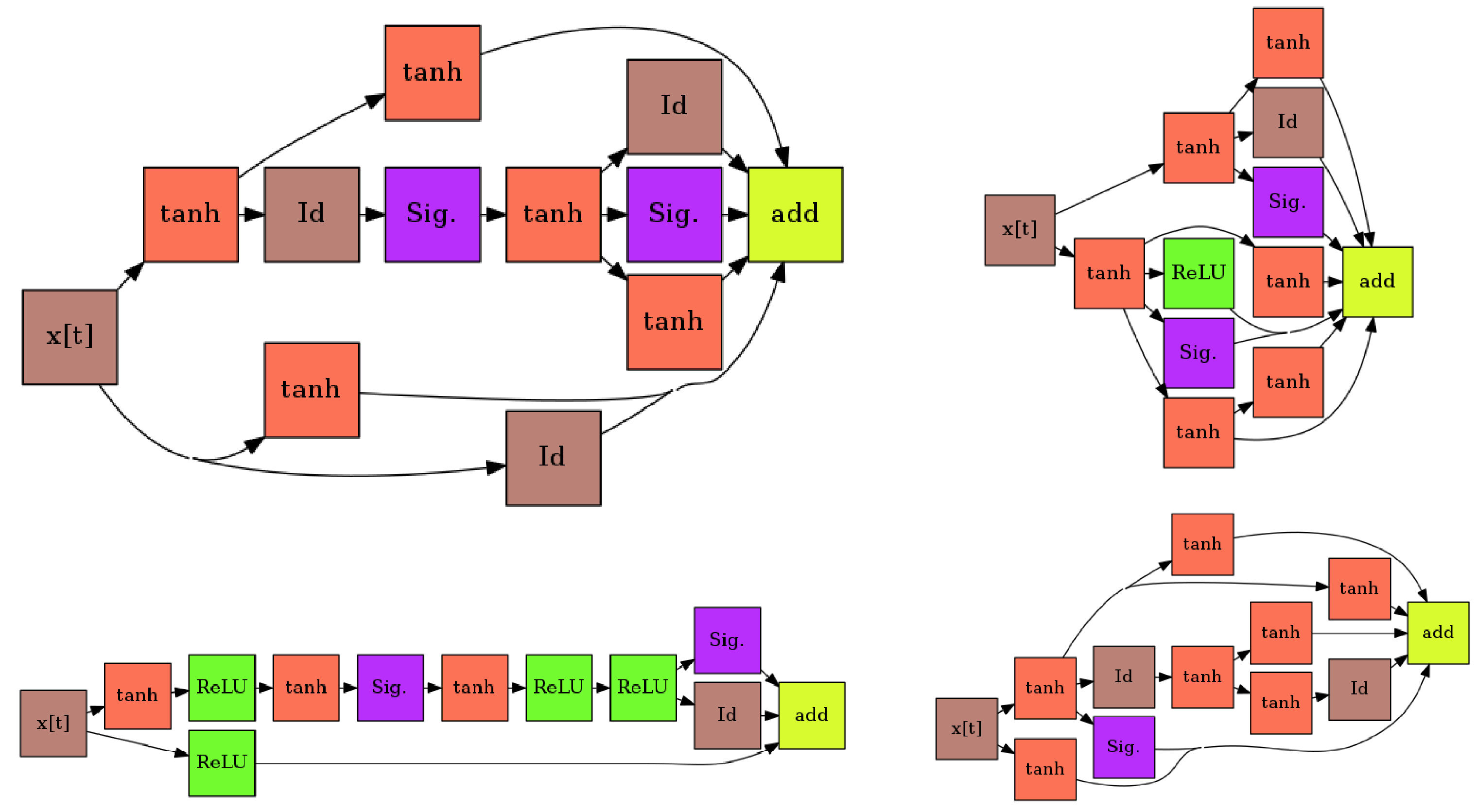}
    
\caption{Few of the recurrent cells used to generate the sequence based models in \textit{ptb} dataset. Each color indicates a different operation from the following: Identity (Id), Sigmoid (Sig.), Tanh (tanh), ReLU (ReLU). $x[t]$ is the input to the cell and the final add operation is its output.
	\label{f:ptb-dataset}
}
\end{figure*}

\subsection{Intermediate Rewards}
\label{s:intermediate-rewards}
\name's MDP reformulation allows us to provide intermediate reward signals that are known to help with the temporal credit assignment problem.
%
%
\begin{figure*}[t!]
\centering
\includegraphics[width=\columnwidth]{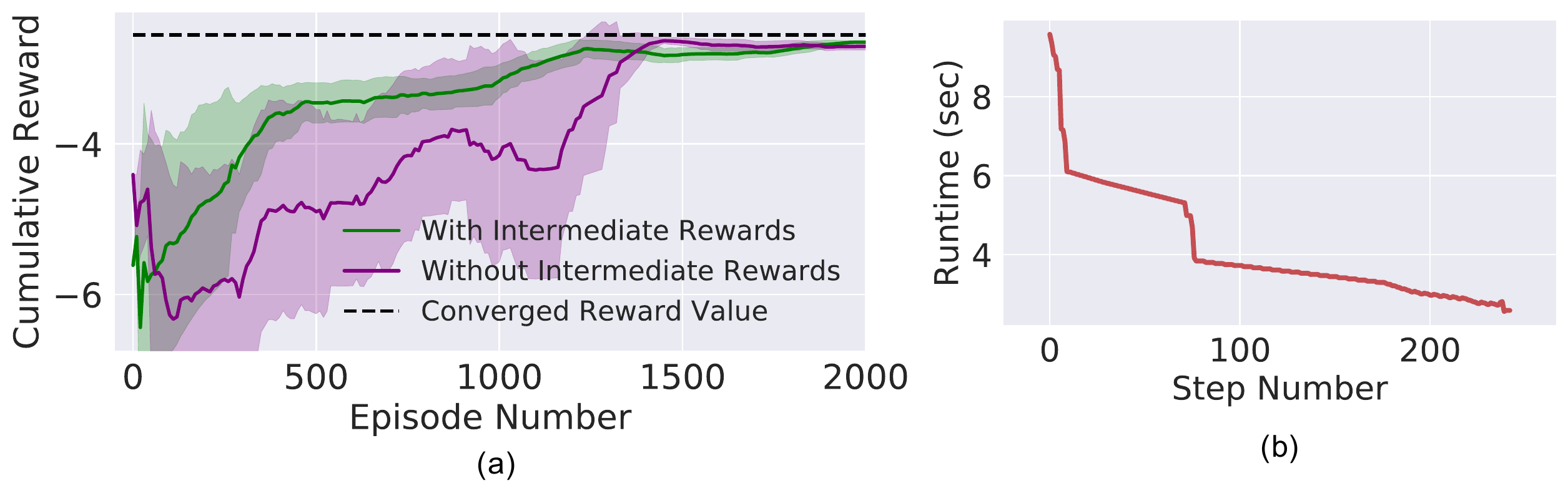}
\vspace{-0.45cm}
\caption{\small
	(a) Cumulative Episodic rewards when \name is trained with and without intermediate rewards on \nmt\, model. Having intermediate rewards within an episode as opposed to a single reward at the end leads to a lower variance in the runtime.
	(b) Runtime improvement observed in the final episode starting from the initial trivial placement.}
\label{fig:int_rew}
\end{figure*}

Figure \ref{fig:int_rew} empirically shows the benefits of having intermediate rewards as opposed to a single reward at the end.
They lead to a faster convergence and a lower variance in cumulative episodic reward terms used by REINFORCE to estimate policy gradients during training.
\name's policy network learns to incrementally generate the whole placement through iterative improvement steps over the course of the episode starting from a trivial placement.

\subsection{Simulator}
\label{s:simulator}

Over the course of training, runtimes for thousands of sampled placements need to be determined before a policy can be trained to converge to a good placement. 
Since it is costly to execute the placements on real hardware and measure the elapsed time for one batch of gradient descent~\cite{grl2, grl1}, we built a simulator that can quickly predict the runtime of any given placement for a given device configuration. 

For any given model to place, our simulator first profiles each operation in its computational graph by measuring the time it takes to run it on all the available devices.
We model the communication
cost between devices as linearly proportional to the size of intermediate
data flow across operations.

The simulator maintains the following two FIFO queues for each device $d$:
\begin{itemize}
	\item $Q^{\text{op}}_d$: \,\, \, Collection of operations that are ready to run on $d$.
	\item $Q^{\text{transfer}}_d$: Collection of output tensors that are ready to be transferred from $d$ to some other device.
\end{itemize}

We deem an operation to be runnable on a device $d$ only after all of its parent operations have finished executing and their corresponding output tensors have been transferred to $d$.

Our simulator uses an event-based design to generate an execution timeline. Each event has a timestamp at which it gets triggered. Further, it also includes some metadata for easy referencing. We define the following types of events:
\begin{itemize}
	\item \textit{Op-done}: Used to indicate when an operation has finished executing. Its timestamp is determined based on the information collected from the initial profiling step on how long it takes to run the operation on its corresponding device.
	\item \textit{Transfer-done}: Used to indicate the finish of an inter-device transfer of an output tensor. Its timestamp is determined using an estimated communication bandwidth $b$ between devices and size of the tensor.
	\item \textit{Wakeup}: Used to signal the wakeup of a device (or a bus) that has been marked as free after its operation queue (or transfer queue) became empty and there was no pending work for it to do.
\end{itemize}

We now define event handlers for each of the above event-types

\textbf{\textit{Op-done} event-handler:}

\hspace{.5cm} Whenever an operation $o$ has completed running on the device $d$, the simulator performs the following actions in order: 
\begin{itemize}
	\item For every child operation $o'$ placed on device $d'$:
		\begin{itemize}
			\item Enqueue output tensor $t_o$ of $o$ to $Q^{\text{transfer}}_d$ if $d \neq d'$.
			\item Check if $o'$ is runnable. If so, enqueue it to $Q_{d'}^{\text{op}}$. 
			\item Add the appropriate \textit{Wakeup} events necessary after the above two steps in case $d'$ happens to be free.
		\end{itemize}
	\item If $Q_d^{\text{op}}$ is empty, then mark the device $d$ as free. Otherwise, pick the next operation from this queue and create its corresponding \textit{Op-done} event.
\end{itemize}

\textbf{\textit{Transfer-done} event-handler:}

\hspace{.5cm} Whenever a tensor $t$ has been transferred from device $d$ to $d'$, the simulator performs following actions in order:
\begin{itemize}
	\item Check if the operation $o$ on device $d'$ that takes $t$ as its input is runnable. If so, enqueue it to $Q_{d'}^{\text{op}}$. Add a \textit{Wakeup} event for device $d'$ if necessary.
	\item If $Q_d^{\text{transfer}}$ is empty, mark the bus corresponding to device $d$ as free. Otherwise, pick the next transfer operation and create its corresponding \textit{Op-done} event.
\end{itemize}

\textbf{\textit{Wakeup} event-handler:}
If a device or its corresponding bus receives a \textit{wakeup} signal, then its corresponding queue should be non-empty. Pick the first element from this queue and create a new \textit{Op-done} or \textit{Transfer-done} event based on it.

We initialize the queues with operations that have no data-dependencies and create their corresponding \textit{Op-done} events. The simulation ends when there are no more events left to process and all the operations have finished executing. The timestamp on the last \textit{Op-done} event is considered to be the simulated runtime.

During simulation, we keep track of the start and end timestamps for each operation. Along with the tensor sizes, these are used to predict the peak memory-utilizations of the devices.

Note that we've tried to model our simulator based on the real execution engine used in Tensorflow. We've validated that the following key aspects of our design match with tensorflow's implementation: (a) Per-device FIFO queues holding runnable operations. (b) Communication overlapping with compute. (c) No more than one operation runs on a device at a time.

\begin{figure*}[!h]
\centering
\includegraphics[width=0.5\columnwidth]{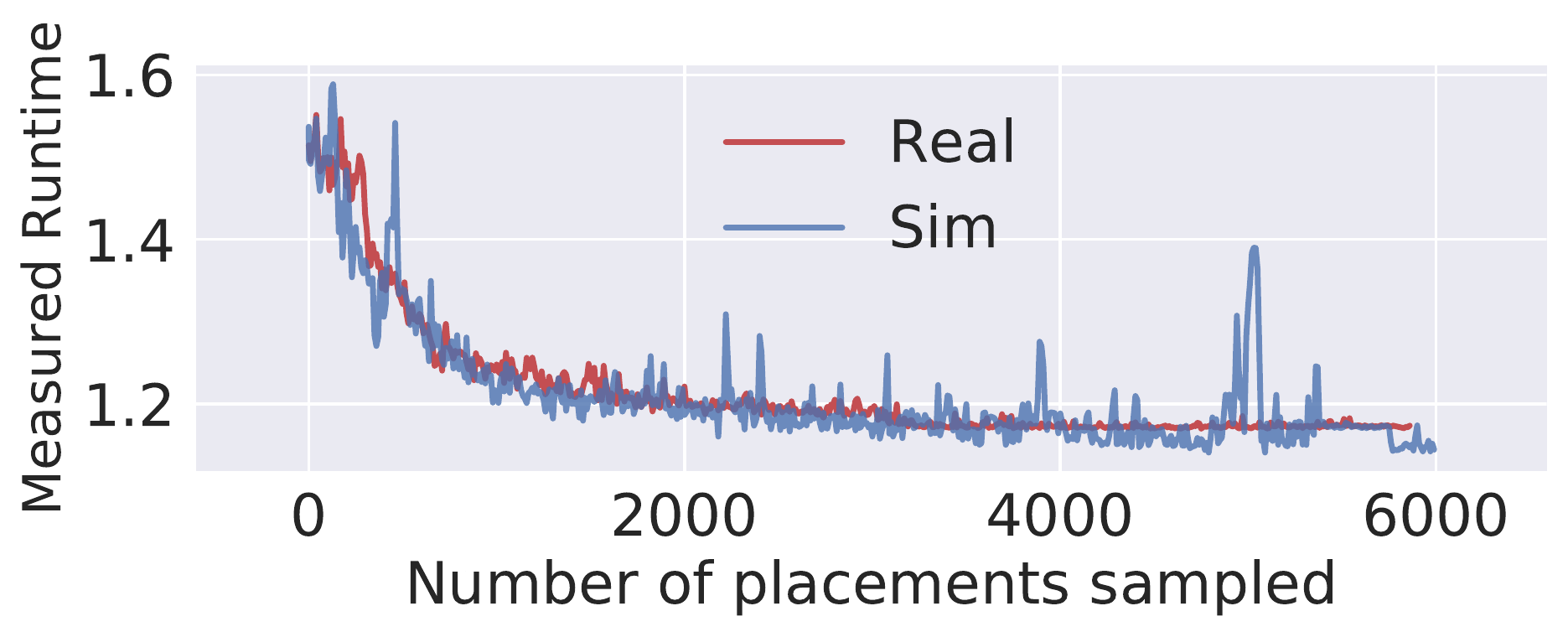}
\caption{\small \rnn-based approach exhibits near identical learning curve when reward signal is from a simulator or directly from measurements on real machines.}
\label{fig:simvsreal}
\end{figure*}

As a result, an RL-based scheme trained with the simulator exhibits nearly identical
run times compared to training directly on the actual system. 
We demonstrate this by comparing the run times in the learning curves of a \rnn-based approach~\cite{grl1} on the real hardware and our simulator~(Figure~\ref{fig:simvsreal}).

\subsection{Merge-and-Colocate heuristic}
\label{s:grouping_merge_strategy}

Merge-and-Colocate is a simple heuristic designed to reduce the size of a graph by colocating small operations with their neighbors.

Given any input graph $G_i$, the Merge-and-Colocate heuristic first merges the node with the lowest cost into its neighbor. If the node has no neighbors, then its predecessor is used instead. 
This step is repeated until the graph size reaches a desired value $N$ or alternatively until there are no more nodes with cost below a certain threshold $C$. 
The merged nodes are then colocated together on to the same device.
For our experiments, we use the size of the output tensor of an operation as the cost metric for the above proceAdure.

\subsection{Training details}
\label{s:training_details}
Here, we describe training details for \name and \rnn-based model. Unless otherwise specified, we use the same described methodology for setting the hyperparameters for both of these approaches.

\textbf{Entropy.} We add an entropy term in the loss function as a way to encourage exploration. 
We tune the entropy factor seperately for \name and \rnn-based model so that the exploration starts off high and decays gradually to a low value towards the final training episodes.

\textbf{Optimization.} We tune the initial learning rate for each of the models that we report the results on. For each model, we decay the learning rate linearly to smooth convergence. We use Adam's optimizer to update our policy weights.

\textbf{Workers.} We use $8$ worker threads and a master coordinator which also serves as a parameter server. At the beginning of every episode, each worker synchronizes its policy weights with the parameter server. 
Each worker then independently performs an episode rollout and collects the rewards for its sampled placement. 
It then computes the gradients of reinforce loss function with respect to all the policy parameters. 
All the workers send their respective gradients to the parameter server which sums them up and updates the parameters to be used for the next episode.

\textbf{Baselines.} 
For \name, we use a seperate moving average baseline for each stage of the episode.
The baseline for time step $t$ is the average of cumulative rewards at step $t$, of the past $k$ episodes where $k$ is a tunable hyperparameter.

For \rnn-based approach, we use baseline as described in Mirhoseini et al.~\cite{grl1}.

\textbf{Neural Network Architecture}
For \name, we use single layer feed-forward networks during message passing and aggregation steps with the same number of hidden units as the input dimension. We feed the outputs of the aggregator into a two layer feed-forward neural network with softmax output layer. We use ReLU as our default activation function.

For the \rnn-based approach, we use a bi-directional \rnn\, with a hidden size of $512$ units.

\textbf{Training Details:}
We use distributed learning with synchronous SGD algorithm to train \name's policy network.
A parameter server is used to co-ordinate updates with 8 worker nodes.
Each worker independently performs an episode rollout and collects the rewards for its sampled placement. 
It then computes the gradients of reinforce loss function with respect to all the policy parameters. 
All the workers then send their respective gradients to the parameter server which sums them up before updating the parameters to be used for the next episode.
To train a policy using multiple graphs, a different graph is used by each worker. More details about the training process including optimization, RL exploration, reward baseline used and neural network architecture descriptions are provided in the Appendix \ref{s:training_details}

\textbf{Reward:}

Given any placement $p$ with runtime $r$ (in seconds) and maximum peak memory utilization $m$ (in GB) across all devices, we define memory penalized runtime, $R(p)$ as follows:
$$ R(p) = 
\begin{cases}
    r                & \text{if } m \leq M\\
    r + c* (m - M)   & \text{otherwise}
\end{cases}
$$
where $M$ is the total available memory on the device with maximum peak memory utilization and $c$ is a scale factor. 
For our experiments, we use $c = 2$.

To find a valid placement that fits without exceeding the memory limit on devices, we include a penalty proportional to the peak memory utilization if it crosses a certain threshold $M$.
This threshold $M$ could be used to control the memory footprint of the placements under execution environments with high memory pressure (\eg GPUs). For instance, we use $M = 10.7$ GB in our experiments to find placements that fit on Tesla K80 GPUs which have about $11$ GB of memory available for use.

For an MDP episode of length T, we propose the following two different ways to assign reward:\vspace{-.25cm}
\begin{itemize}[leftmargin=.8cm]
	\item \textbf{Terminal Reward:} A Non-zero reward is given only at the end of the episode. 
	That is, $r_1=0, r_2=0, \hdots r_T=-R(p_T)$. This requires evaluating only one placement per episode but leads to a high variance in the policy gradient estimates due to a lack of temporal credit assignment.
	\item \textbf{Intermediate Rewards:} Under this setting, the improvement in runtimes of the successive time steps of an episode is used as an intermediate reward signal. 
	That is, $r_1=R(p_1)-R(p_0),\quad r_2=R(p_2)-R(p_1),\quad \hdots\quad r_T=R(p_T)-R(p_{T-1})$. Although this requires evaluating\, $T+1$ placements for every episode, intermediate rewards result in better convergence properties in RL \cite{Williams92simplestatistical}.
\end{itemize}

\textbf{Devices:}

We target the following device configuration for optimizing placements: 
Tensorflow r1.9 version running on a p2.8xlarge instance from AWS EC2 \cite{aws_instance_page} equipped with $8$ Nvidia Tesla K80 GPUs and a Xeon E5-2686 broadwell processor.

\end{document}